# Mapping and Describing Geospatial Data to Generalize Complex Models: The Case of LittoSIM-GEN

## Abstract

Simulation models respond to scientific questions based on internal calculations using input data. These data usually come from different sources with diverse and heterogeneous formats. The design of complex data-driven models is often shaped by the structure of the data available in research projects. Hence, applying such models to other case studies requires either to get similar data or to transform new data to fit the model inputs. We faced this problem in the LittoSIM-GEN project when generalizing our participatory flooding model (LittoSIM) to new territories. From this experience, we provide a mapping approach to structure, describe, and automatize the integration of geospatial data into complex models.

## 1 Introduction

Nowadays, scientific models are dealing with real data to approach reality and to simulate studied phenomena more efficiently [12, 27]. The availability of big data and the expansion of calculating machines foster this tendency, particularly in social sciences, where the research is progressively empirical and data-driven [2, 25, 12]. The development of complex models with large amounts of data makes them more complicated to understand, replicate, and reuse [21, 16, 20], especially when it comes to generalizing them to new data [22, 3, 15]. This task becomes more challenging when these data are noisy, heterogeneous, or incomplete.

Transformation, aggregation, cleaning, and other processing operations are often required to adapt and link separate data sources to models [15]. The heterogeneity of data structures makes manual processing time and effort consuming [9]. This difficulty hinders reusability (reproducibility and replicability [8]) and pushes modelers to develop new models from scratch instead of reusing existing models with similar purposes [22, 17]. Others tend to omit or to reduce the amount of considered empirical data, but this leads to theoretical or toy models that may be inappropriate for applied modeling projects [2, 25]. Therefore, scientists need approaches, methods, and tools to simplify data preparation and integration through formalized and automatized procedures.

The absence of tools that describe the data-model relationship adds more complications to data integration and decreases the reusability of models [17, 15, 9]. As a result, data preparation is hard to reproduce for other contexts where datasets are different. Hence, in addition to methods that process and prepare data, researchers need to trace their processing by describing data, models, and data-model mapping [20, 15]. Sharing data is common in some disciplines [13]. However, it is not always possible or enough to replicate or reproduce models when not all relevant information is shared [21, 17]. Therefore, standard mapping and description protocols are also needed to make reusing models with new data more convenient [15, 18].

In this work, we argue that data mapping and describing are necessary to generalize and reuse [6] complex models. According to our experience with the LittoSIM-GEN project, reusing existing models with different data is challenging in geospatial modeling. Operations such as transformations, intersections, and topological verification are always required to reuse models with new data. Mapping schemes and textual descriptions document this process and help to develop data-compilers. These automated scripts prepare data for model inputs and reduce the amount of time and effort required for data integration. Sharing these data-compilers with standard descriptions increases the reusability of complex geospatial models. We demonstrate this approach through the case of LittoSIM [4], a model of coastal flooding risk learning and management.

This paper is structured as follows. The next section presents data-related generalization issues of existing geospatial models. In the third section, we present the LittoSIM model with its data structure, and then we introduce the LittoSIM-GEN project. The fourth section presents the mapping approach used to describe and process data to address the genericity issue of the model. Then, we apply this approach to the LittoSIM-GEN case. Finally, we conclude with some conclusions and perspectives of this work.

## 2 Existing geospatial models and their generalization issues

Geospatial models are often designated to address real-world problems by studying natural and social complex phenomena. Developed models are intended to discover hidden patterns and extract valuable knowledge that helps decisionmakers. This realistic tendency constrains these models to use big amounts of real data, which may increase their efficiency and validity, but it also their complexity that raises serious issues for geographers and modelers when trying to reuse existing models. Reusing a model with new data is not straightforward when how to integrate it is not specified, particularly in geospatial modeling where data are multidimensional [31] and heterogeneous [21]. Agent-based models, for example, are conceived to allow the modeling of the diversity in geospatial data and their evolution over time [36], but their coupling is not evident and requires the specification of representations and relationships between GIS data and ABMs [31].

In the domain of coastal risk management, efforts are made to develop more generic decision support tools [32, 33]. For instance, authors of [35] use GIS-based methods to provide decisionmakers with information such as the length of time needed for an area to become over-flowed by a particular storm surge level. The issues are about combining multiple data types (geographic, statistical, …) and integrating different data domains (hydrodynamic, spatial planning, social perception, …) [34]. This need for generalization is urged by the development of new modeling techniques, and by the need to learn new risk prevention strategies.

In the last few years, simulations and serious games became a concrete tool that uses participatory modeling to foster learning and improve awareness in an attractive way [29]. Multiple research works have investigated the use of serious games among different environmental disciplines. However, in coastal flooding management, besides SPRITE [30] designed for students, we found only LittoSIM [4] as a serious game implying decisionmakers.

# 3 LittoSIM: a participatory simulation model

LittoSIM [4] is a project that aims at modeling the effects of coastal flooding on inhabited areas and the impact of land use management and defense measures on the extent of flood events. LittoSIM uses a policy game [7] to make decisionmakers experience alternative strategies for decreasing potential damage [28], as flooding costs may depend on individual decisions and risk awareness [10]. Participants in LittoSIM control the simulation based on their expectations and objectives according to a set of constraints that constitutes the game rules.

## 3.1 LittoSIM model

LittoSIM is implemented as a participatory computer game [24] through a set of agent-based models under the GAMA platform [26]. The graphical interface of LittoSIM uses the features of GAMA. Network communications use the Apache ActiveMQ wrapper[1] to run the game between different participants over the network. The flooding module integrates a 2D hydrodynamical model called LISFLOOD-FP [19]. Figure 1 depicts the global architecture of LittoSIM.

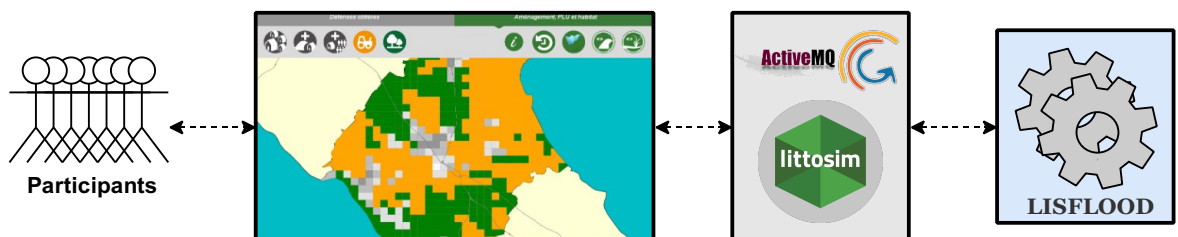

*Figure 1:  The architecture of LittoSIM. Participants manipulate GAMA graphical interfaces to interact with the LISFLOOD model and with other participants over the network through ActiveMQ.*

The participants in a LittoSIM game are of three types. (i) The players representing the districts of the study area, they manage their territories through actions of land use planning and coastal defenses. (ii) The game leader represents the role of the state risk agency, he provides policy advice to players and manages meetings between them to elaborate collective strategies. (iii) The game manager manages game rounds and launches submersion events. The flooding results are calculated by LISFLOOD based on two configuration files (bdy for the event scenario and bci for the event location [19]) and the current state of the study area. This state is represented by two grids of soil altitudes (DEM: Digital Elevation Model) and soil resistance to water (rugosity). The result is returned to GAMA as a set of grids of water elevation over the study area (figure 2).

---

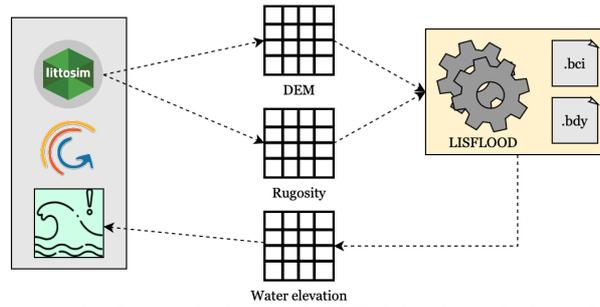

*Figure 2: The grid exchange between the GAMA platform and LISFLOOD that calculates the flood propagation based on two configuration files: bdy (submersion scenario) and bci (submersion location).*

LittoSIM was initially developed for the districts of Oléron island [4] on the French Atlantic coast. This type of models dealing with natural disasters has to be knowledge-based and data-driven [23]. Hence, the usage of real data was primordial and shaped the conception of the model [1].

## 3.2 LittoSIM data

LittoSIM is a data-specific model that needs predefined data files and formats to run correctly. These files depicted in figure 3 represent different spatial aspects of the study area that are deemed necessary to approach the problem of coastal flooding management.

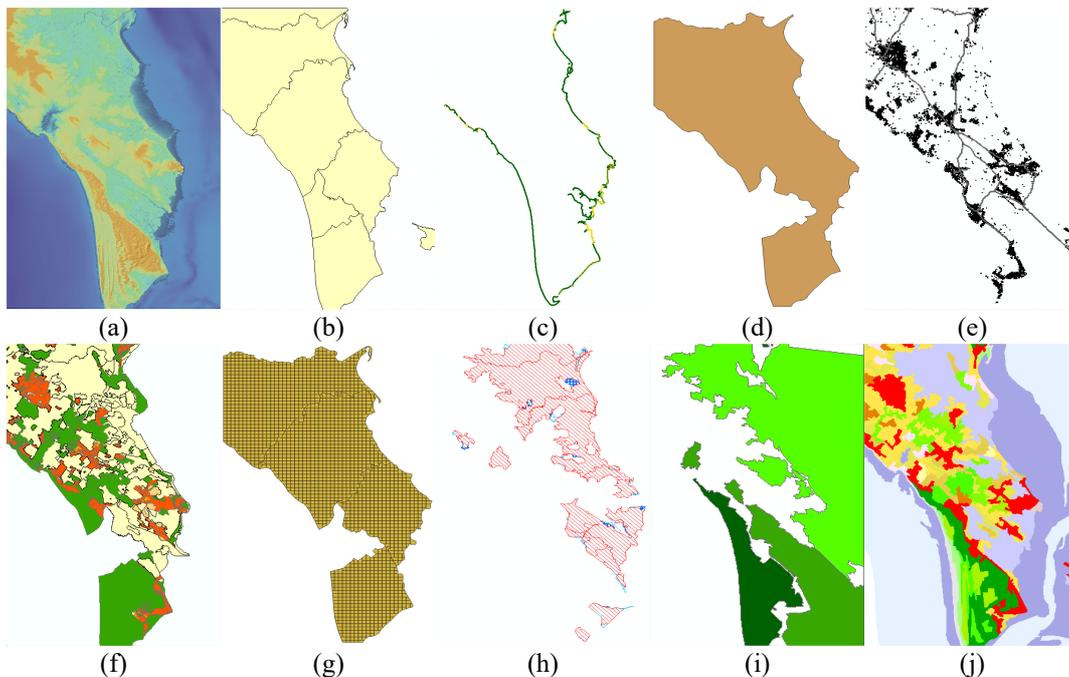

*Figure 3: Areas and spatial components of LittoSIM-Oléron: (a) digital elevation model; (b) districts; (c) coastal defenses; (d) inland dikes area; (e) buildings; (f) local urban planning; (g) land use grid; (h) risked areas; (i) natural protected areas; (j) land cover.*

Table 1 lists all the necessary files to run a LittoSIM game. Files marked as <u>primary</u> must be downloaded from dedicated geospatial databases to provide real relevant data (*districts*, *urban_plan*, *land_cover*, *buildings*, *coastal_defenses*, *coastline*, *spa*, *rpp*, *dem*, and *file.bdy*). Some files (*roads*, *water*) are optional and can be omitted as they serve only for visualization purposes.

*Table 1: LittoSIM input files. The parenthesized letters refer to sub-figures in figure 3.*

| File | Type | Nature | Description |
|---|---|---|---|
| *districts* | shapefile | primary | the districts of the study area **(b)** |
| *convex_hull* | shapefile | secondary | a rectangular buffer enveloping the study area |
| *buffer_in_100m* | shapefile | secondary | a buffer of 100m inside the districts' shape **(d)** |

| *land_use* | shapefile | secondary | a grid of land use and planning cells **(g)** |
|---|---|---|---|
| *urban_plan* | shapefile | primary | the local urban plan (PLU: Plan Local d'Urbanisme) specifying the town planning **(f)** |
| *land_cover* | shapefile | primary | the Corine Land Cover (CLC) specifying the land cover type **(j)** |
| *buildings* | shapefile | primary | the buildings of the study area **(e)** |
| *roads* | shapefile | primary | lines representing principal roads (optional) |
| *water* | shapefile | primary | lines representing principal rivers (optional) |
| *coastline* | shapefile | primary | a line representing the coast |
| *coastal_defenses* | shapefile | primary | dunes and dikes protecting the coast **(c)** |
| *spa* | shapefile | primary | natural and special protected areas **(i)** |
| *rpp* | shapefile | primary | the risk prevention plan of flooding risk areas **(h)** |
| *dem* | raster | primary | the digital elevation model (soil altitude) grid **(a)** |
| *rugosity* | raster | secondary | soil roughness (resistance to water) grid |
| *file.bdy* | text | primary | a time series of the water elevation scenario |
| *file.bci* | text | secondary | geographical boundaries of the domain |

The other files marked as <u>secondary</u> in table 1 (*convex_hull*, *buffer_in_100m*, *land_use*, *rugosity*, *and file.bci*) are created based on primary shapefiles using GIS and data processing techniques as follows (the rugosity grid uses the Manning roughness [4] coefficients based on Corine Land Cover data):

$$districts \rightarrow convex\_hull$$

$$districts \rightarrow buffer\_in\_100m$$

$$districts \cap convex\_hull \rightarrow file.bci$$

$$land\_cover \cap \text{Manning coefficients} \rightarrow rugosity$$

$$districts \cap urban\_plan \cap land\_cover \cap buildings \rightarrow land\_use$$

These processing operations are not straightforward and various spatial transformations are required to generate the appropriate files for LittoSIM. Figure 4 shows two examples of merging and splitting polygons while creating the land use grid. In addition, all previous shapefiles have to be validated by fixing geometrical (*e.g.* overlapping polygons) and topological (*e.g.* dangling nodes) errors. Projection and coordinate systems have also to be the same for all files. The LittoSIM model requires also these files to have a specified data structure as depicted in table 2 (all shapefiles have an additional automatic ID of type integer).

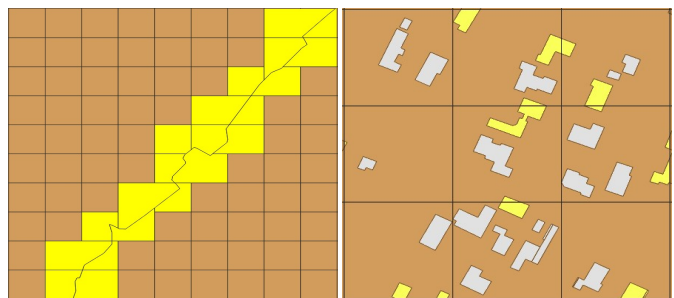

*Figure 4: Merging border cells and splitting inter-cellular buildings.*

For flooding simulation, the submersion data provides a time series file (*file.bdy*) defining water elevations for several landmarks (geographical boundaries) over the study area. These landmarks are specified in a second file (*file.bci*) by determining the coordinates of selected points. These two files, along with the two raster grids (*dem* and *rugosity*), are required to launch the LISFLOOD model.

Hence, applying LittoSIM to new case studies requires transforming all available data into the previous textual and spatial structures. Simplifying this task is the main purpose of the LittoSIM-GEN project.

*Table 2: The structure of LittoSIM input data. All files have a default attribute ID of type integer.*

| File | Attribute | Type | Values | Description |
|------|-----------|------|--------|-------------|
| *districts* | *player_id* | Integer | [1,4] | identifies active districts |
| | *dist_code* | String | | unique code of the district |
| | *dist_sname* | String | | short name of the district |
| | *dist_lname* | String | | full name of the district |
| | *dist_pop* | Integer | | total population |
| | *dist_area* | Double | | total area (m$^2$) |
| *buildings* | *bld_type* | String | *{Residential, Other}* | type of the building |
| | *bld_area* | Double | | the building area (m$^2$) |
| *urban_plan* | *unit_code* | Integer | [1,5] | PLU type of the area |
| *land_cover* | *cover_type* | Integer | {111, …, 523} | CLC type (44 classes) |
| *coastal_defenses* | *dist_code* | String | | district code of the object |
| | *type* | String | *{Dike, Dune}* | type of the object |
| | *status* | String | *{Good, Medium, Bad}* | status of the object |
| | *alt* | Double | | the altitude of the object |
| | *height* | Double | | height of the object |
| *land_use* | *unit_code* | Integer | [1,5] | type of the cell |
| | *sub_type* | Integer | | sub-type of the cell |
| | *unit_pop* | Integer | | the population of the cell |
| | *dist_code* | String | | district code of the cell |
| | *unit_area* | Double | | area of the cell (m$^2$) |
| | *expro_cost* | Double | | expropriation cost of the cell |

## 3.3 LittoSIM-GEN project

LittoSIM-GEN is a project that aims at extending the LittoSIM model to many territories in metropolitan France (Oléron, Normandie, Camargue, Rochefort, ...). Each use case has its specific data with different files and formats. Figure 5 shows, for example, the structure of original files representing the urban local plan (*urban_plan*) of Oléron and Normandie. Only one attribute is relevant for LittoSIM: *unit_code* (table 2). This information comes from two files with different names and structures. Preparing data in LittoSIM-GEN consists of processing sources to create the target file used by the LittoSIM model. This file must have the attribute coded as an integer of four values ({1,2,4,5}) based on the heterogeneous sources ({*A, AU, N, U*} and {*A, AU, AUc, AUs, N, Nh, U*}).

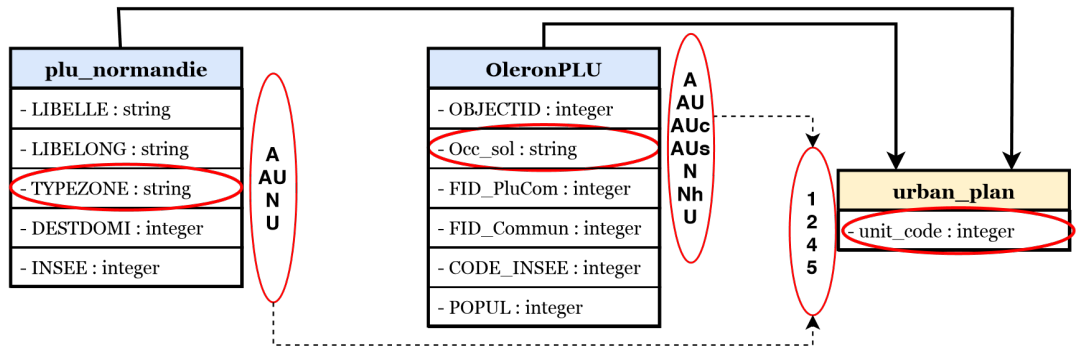

*Figure 5: Blue entities are the Oléron and Normandie files that represent the local urban plan. These files with different structures are mapped to the LittoSIM-GEN file "urban_plan".*

Besides complex spatial processing (figure 4), additional operations of textual conversion and replacement are therefore needed to meet LittoSIM requirements in terms of data structure (table 2). Removing unused fields is also important to avoid big files with useless data. Hence, applying LittoSIM-GEN to new territories where data are different requires: (i) understanding the data structure of LittoSIM; and (ii) transforming local data into this structure. Therefore, an approach of data mapping, describing, and processing is needed to make LittoSIM-GEN a generic tool easy to reuse by third parties.

# 4 Mapping, describing, and processing data

Data mapping consists of defining necessary operations to transform source data into the target structure required by a model [14] (e.g. LittoSIM). A mapping scheme allows telling where relevant information comes from and how it is used in the model. Such data workflow is often complicated and hard to disseminate to others. Therefore, it must be described to understand how to reproduce its results. For this task, we use an extension of the ODD (Overview, Design concepts, and Details) protocol [11] called ODD+2D (ODD + Decision + Data) [15]. It focuses on the input data structure to describe the processing and integration of raw data into an agent-based model.

In this section, we present our data mapping approach that consists of three processing steps. The structure of ODD+2D serves as a container and allows us to document and link these steps through the use of mapping patterns. An automatized data-compiler is then implemented to process data based on the described mapping.

## 4.1 A three-steps mapping approach

The mapping process experienced with LittoSIM-GEN is divided into three steps (figure 6). The first step consists of identifying the original data to determine the <u>source structure</u>. Modelers can select only the data used in the model. However, it may be relevant to describe all related data sources to understand the context of modeling and data integration [12, 15]. In the second step, a <u>target structure</u> is defined to respond to the research question. Target data are the set of files required to feed model inputs and simulation parameters (*e.g.* LittoSIM data structure of tables 1 and 2). The last step consists of bridging the gap between the two structures by determining different <u>mapping patterns</u> and their underlying processing operations such as transformation, aggregation, and generation.

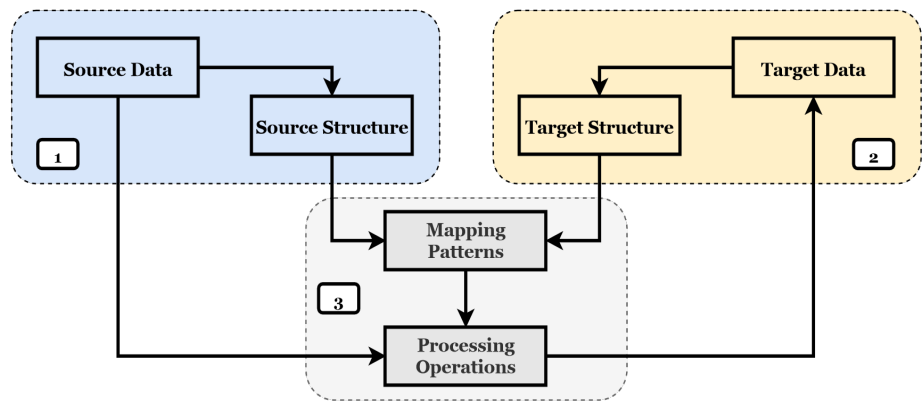

*Figure 6: Data mapping process in three steps: source identification, target identification, and linking the two structures through mapping patterns.*

## 4.2 Geospatial mapping patterns and ODD+2D

In geospatial modeling, several mapping patterns are needed to process spatial and textual features. Figure 7 depicts the patterns used in LittoSIM-GEN. Each pattern uses one (1) or many (0..*: many, 1..*: at least one, 2..*: at least two) sources to create the target attribute/file.

**Attribute patterns**: these patterns are the basic textual transformations between attributes such as *Rename* for renaming, *Convert* for type conversion, and *Shorten* for attribute size shortening. These patterns can be aggregated to form one complex pattern (*e.g. Rename+Convert*). Five more attribute patterns are used. Parenthesized examples refer to LittoSIM data of table 2.

➢ *Replace*: replaces the values of an attribute (*e.g. coastal_defenses.status*).

➢ *Aggregate*: combines several (2..*) attributes into a new one (*e.g. land_use.unit_pop*).

➢ *Generate*: creates a new attribute that does not exist in the original data (*e.g. districts.player_id*).

➢ *Intersect*: creates new spatial objects that result from an intersection between two or more geometries. In parallel, this pattern generates new attributes by superposing textual features of the intersected objects. It requires at least two source files (*e.g. land_use.dist_code*).

➢ *Merge*: merges two or more geometries to form a new one. The pattern should determine how attributes of the new object are created based on the merged data tables. This pattern requires at least two sources (*e.g. land_use.unit_area*).

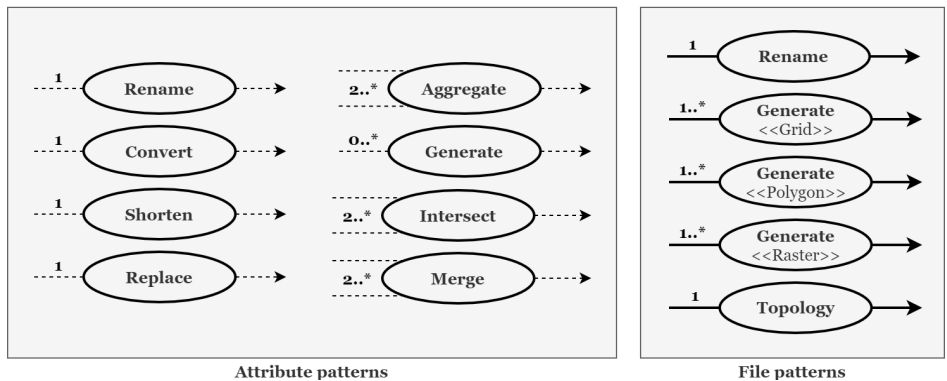

*Figure 7: Geospatial data mapping patterns. Attribute patterns are linked with dashed lines. File patterns are linked with solid lines. Multiplicities indicate the number of elements required for each pattern.*

**File patterns**: a file can be just copied and renamed (*Rename*). However, more processing operations may be necessary to transform or to create new spatial features in the target data. The processing script reads the configuration file and transforms data into the required structure. Parenthesized examples refer to LittoSIM data of table 1.

- *Rename*: copies and renames a file (*e.g. communes → districts*).

- *Generate*: creates a new file based on the existing source files. It concerns mainly generating grid cells <<*Grid*>> (*e.g. land_use*), raster grids <<*Raster*>> (*e.g. rugosity*), and spatial geometries <<*Polygon*>> (*e.g. convex_hull*).

- *Topology*: verifies the processed files by fixing geometrical and topological errors. It is required to verify generated files through splitting and merging polygons to avoid erroneous manipulation of spatial features.

## 4.3 Mapping patterns to XML-R data-compiler

The previous mapping patterns are translated into an XML code that configures an R processing script to generate target data. Each case study has its XML configuration file specifying source and target structures, and the set of transformations to make. The processing script reads the configuration file and transforms data into the required structure. Choosing tools and techniques to implement the script depends on the nature of the data and the needed transformations. In the case of LittoSIM-GEN, we use R as it is an open-source solution offering all necessary packages to process geospatial data. Figure 8 shows the architecture of the XML-R data-compiler used in the LittoSIM-GEN project.

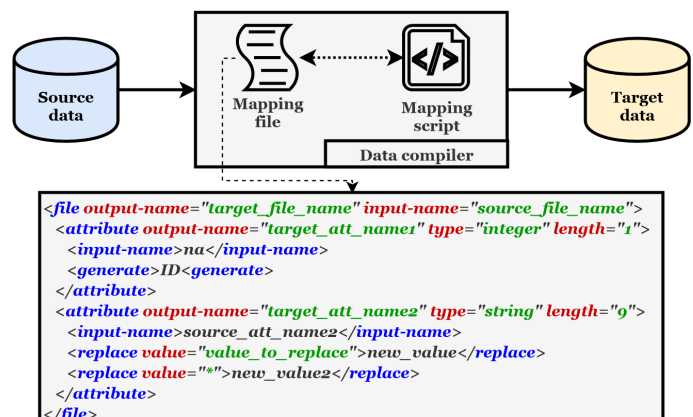

*Figure 8: The architecture of the XML-R data-compiler. The XML mapping file stores source and target structures with the mapping patterns between them. This file feeds the R script that processes data.*

# 5 Application to LittoSIM-GEN

In this section, we present the previous mapping approach as applied in the case of LittoSIM-GEN. We provide the ODD+2D description of data, then we implement the LittoSIM-GEN data-compiler.

## 5.1 ODD+2D description of LittoSIM-GEN data

The following description is based on the Oléron island case study [4]. We describe only the "Input data" element of the protocol, and we omit other parts, as describing agents and system dynamics is not the focus of this paper.

**Data overview**

Data represent the districts of Oléron island in the Bay of Biscay. These data are gathered from various sources. Administrative and topographical data correspond to the IGN (Institut Géographique National) BD-TOPO® database. Land use data are from the European Union Corine Land Cover (2012). Local urban planning data are provided by local authorities. Protected natural areas are from the local French State. Storm, coastal defenses, and risk prevention plan files come from the research program RISKS (2014) and the state department database DDTM (17-2015). Additional coastal defenses data (position, height, length, type, state) were manually collected. LISFLOOD data scenarios are based on the Xynthia storm [5].

**Data structure**

Table 2 represents the internal structure (attributes and their corresponding values) of data required by the LittoSIM model (files listed in table 1). In addition to these attributes, each file has a default auto-generated numeric ID. The mapping scheme in figure 9 represents the original structure of Oléron data with the entities in blue. The target data structure is in orange.

**Data mapping**

Mapping patterns (gray ovals) depicted in figure 9 summarize data transformations to the LittoSIM structure. All primary files are copied from sources, and an automatic ID is created for each spatial object of the target shapefiles. Unused attributes and features are removed. Basic patterns are self-explaining (e.g. *Rename* is a simple operation of copying and renaming the original data). Complex transformation patterns are identified by their output name (file or attribute) and are explained in the next section (data patterns).

**Data patterns**

❖ *districts*: represents the districts of the study area (island of Oléron).
  ➢ *player_id*: is created to specify the playing districts that take numbers from 1 to 4 (for performance purposes, players in a LittoSIM game are limited to 4). Other districts take a value of 0.
  ➢ *dist_sname*: a short name (10 characters max) is created for each district for identification, simplification, and visualization purposes.
  ➢ *dist_area*: generates the surface of a district by calculating the area of its polygon.

❖ *buffer_in_100*: represents a bounding zone of 100 meters inside the unified districts' area. It is used to identify a coastal defense (dike) as littoral or inland (retro-dike).

❖ *convex_hull*: generates a convex hull specifying the global study area for the simulation as a rectangular closure (envelope) around the districts.

❖ *buildings*: represents the buildings of the study area.
  ➢ *bld_type*: as we are interested only in residential buildings, this attribute contains two values ({*Residential*, *Other*}) created by a pattern of replacement by replacing "*résidentiel*" with "*Residential*", and all other values "*\**" with "*Other*".
  ➢ *bld_area*: calculates the surface of the building based on its polygon.
  ➢ *Topology*: the topological verification of buildings file consists of splitting polygons intersecting with several land use cells, and deleting buildings located in natural areas.

❖ *urban_plan*: represents the land use planning (Plan Local d'Urbanisme) that determines the exploitation type of local areas.
  ➢ *unit_code*: specifies the type of each area in urban planning based on the original file with a type conversion (text to numeric) and replacement of values as follows:
    {N, Nh} → 1, {U, Us} → 2, {AUc, AU, AUs} → 4, A → 5

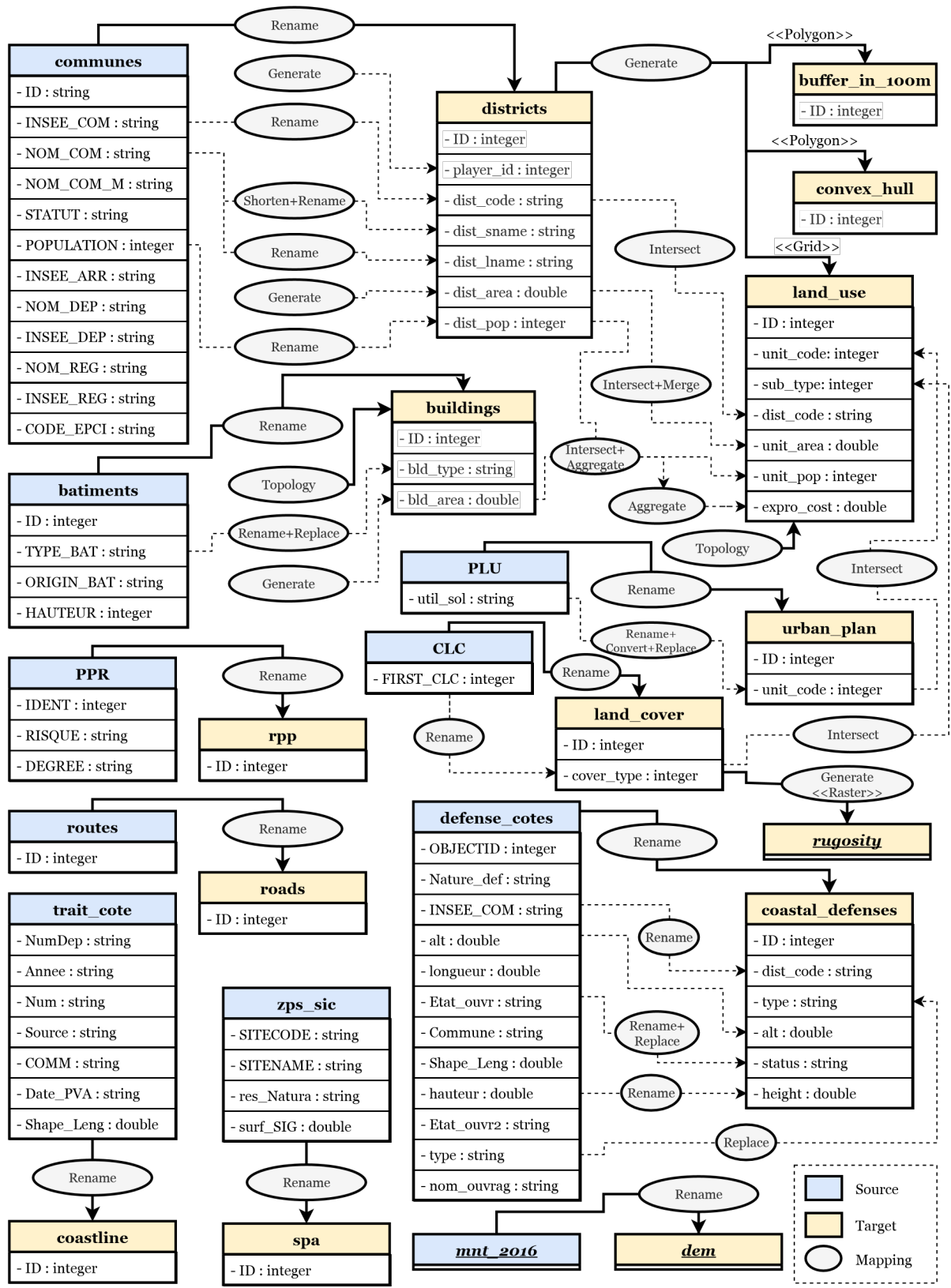

*Figure 9: Mapping diagram of LittoSIM data. Blue entities represent the source data. Orange entities represent the target data. Solid lines (—) link file mapping patterns. Dashed lines (---) link attribute mapping patterns.*

❖ *land_cover*: determines the land cover type of the area based on the European Corine Land Cover (CLC) database.
  ➢ *cover_type*: specifies the type of each zone in the study area. This attribute can take any of 44 standard values ({111, ..., 242, ..., 523}) representing various land covers (forests, airports, rice fields, ...).

❖ *land_use*: is generated as a grid cell of a specified size (200x200m for Oléron) covering the area of the active districts. This grid is the principal component allowing planning and land use management (through editing cells) in LittoSIM.

   ➢ *unit_code*: is created through an intersection between the land use unit (one cell of land use) and the urban planning file (urban plan). It specifies the planning type of the cell and takes initially four values: 1 (Natural), 2 (Urbanized), 4 (Authorized for urbanization), and 5 (Agricultural). During the game, this variable may take additional values: 6 (Urbanized special), 7 (Authorized for urbanization special).

   ➢ *sub_type*: if it applies, this attribute specifies the sub-type of a land use unit. For example, in some cases, agricultural areas may have various sub-types. It is created by an intersection with the land cover file.

   ➢ *dist_code*: affects the district code to each land use cell through an intersection with the districts file. Border cells are split to fit within districts' boundaries.

   ➢ *unit_area*: calculates the surface of each *land_use* unit. For ergonomic purposes (players cannot select miniature cells), a cell smaller than a specified minimum size (20.000 m$^2$ for Oléron) is merged with the neighbor that shares the longest border with it.

   ➢ *unit_pop*: uses aggregate and intersect patterns to calculate the population of each land use unit. It takes the sum of the area of residential buildings intersecting with the cell (*bld_area*), divides it by the sum of the area of residential buildings of the district, and multiplies the result by the total population of the district (*dist_pop*). LittoSIM considers that the population rate is equal to the proportion of residential buildings within each cell (equation 1).

$$resid\_blds \leftarrow buildings\ [bld\_type = "Residential"]$$
$$unit\_pop \leftarrow \frac{\sum(bld\_area,\ resid\_blds \cap land\_use)}{\sum(bld\_area,\ resid\_blds \cap districts)} * dist\_pop$$

*Equation 1: Calculating the population of each unit cell by aggregating multiple variables.*

   ➢ *expro_cost*: the expropriation (converting to natural) cost of an urban cell depends on its population. Empty cells (*unit_pop* = 0) take a predefined parameter (*empty_expro_cost*) in LittoSIM. Populated cells are expropriated with a cost that follows the next function (equation 2). This cost increases less as the cell population goes up. The number 400 represents the expropriation cost of a cell with one inhabitant.

$$expro\_cost \leftarrow \frac{unit\_pop * 400}{\sqrt{unit\_pop}}$$

*Equation 2: Calculating the expropriation cost of the unit based on its population.*

   ➢ *Topology*: the topological verification of the land use grid consists of fixing cell merging issues and avoiding that a natural cell contains buildings or that an agricultural cell has populations (agricultural cells may contain nonresidential buildings).

❖ coastal defenses: represents the dunes and the dikes that protect the coastline of districts from flooding risks.

   ➢ *type*: specifies the type of coastal defense among the two possible values. This attribute is created with a replacement of original values: "*Naturel*" becomes "*Dune*" and all other values "*\**" become "*Dike*".

   ➢ *status*: determines the quality of a coastal defense that affects its probability of rupture during flood events. This attribute is renamed and replaces old values as follows:
   "*bon*" → "*Good*", "*moyen*" → "*Medium*", "*mauvais*" → "*Bad*"

❖ rugosity: the rugosity grid has the same dimensions as the dem grid (631x906 of 20x20m cells for Oléron). It is created based on a predefined set of rugosity coefficients (Manning) depending on the land cover of the area covering each cell. The initial rugosity grid is generated with the detailed land cover (CLC) data. During the game, updated cells take simplified coefficients corresponding to the land use cell type. In the case of Oléron, we use the following coefficients: Natural N (0.11), Urbanized U (0.05), Authorized for urbanization AU (0.09), Agricultural A (0.07), Urbanized special Us (0.09), Authorized for Urbanization special AUs (0.09). Therefore, each updated standard land use cell (200x200m) will contain 100 rugosity cells (20x20m) with the same coefficient.

## 5.2 The LittoSIM-GEN data-compiler

The previous mapping and description explain how to get the LittoSIM structure from Oléron data. Such a process can be applied to any data structure belonging to another territory. Transformation patterns are translated into the configuration XML file by specifying the source and the target data, along with the mapping operations between them. For example, the next XML code shows an excerpt from the Oléron mapping file: the attribute *dist_code* of the file *districts* is created as a String (5) based on the *INSEE_COM* attribute of the original *communes* file.

```xml
<file output-name = "districts.shp" input-name = "communes.shp">
    <attribute output-name = "dist_code" type = "string" length = "5">
        <input-name>INSEE_COM</input-name>
    </attribute>
    <attribute output-name = "dist_lname" type = "string" length = "40">
        <input-name>NOM_COM</input-name>
    </attribute>
    >
    ...
</file>
```

This mapping XML file is then read by a mapping script written in the R language. The script performs all the data processing operations by transforming files/attributes, editing spatial objects, correcting projections, and generating missing data as specified in the configuration file. Fixing standard topological errors is done manually through GIS software, as processing such errors is difficult to automatize. However, geometric cases considered as errors by LittoSIM (population in natural cells, inter-cellular buildings) are handled by the data-compiler. The next R code is an excerpt from the processing script that fixes projections and generates an integer ID for spatial objects. These two files (XML + R) compose the LittoSIM-GEN data-compiler.

```r
# reprojecting shapefiles to Lambert-93 (<<Topology>>)
if(is.na(proj4string(shapefile))){
        proj4string(shapefile) <-
        CRS("+init=epsg:2154");
} else {
        shapefile <- spTransform(shapefile, CRS("+init=epsg:2154"));
}
# creating an ID for spatial objects (<<Generate>>)
shapefile$ID <- 1:length(shapefile);
...
```

The full source code of the LittoSIM-GEN data-compiler (XML and R files) is given in appendix A. This tool allows automatizing the process of LittoSIM data preparation. In addition to time and effort saving, the generated files are more accurate compared to those processed manually using GIS software. A comparison example is depicted in figure 10. Grid cells fit better districts' boundaries with split and merge operations. This accuracy affects the game quality by improving visualization and avoiding erroneous manipulations of spatial objects.

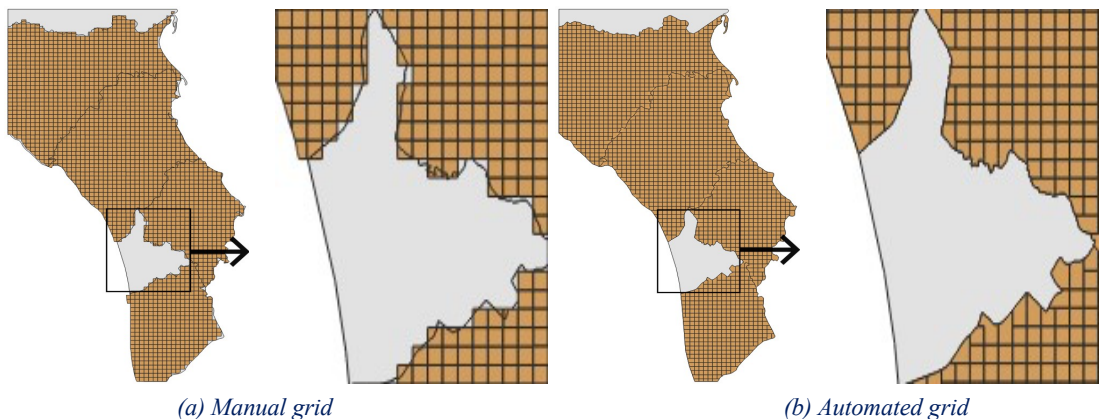

(a) Manual grid            (b) Automated grid

Figure 10: Manual and automated processing of the land use grid

# 6 Conclusions and perspectives

In this paper, we presented a process to enable the generalization of a coastal flooding risk simulation model. It uses data mapping and describing as tools to ease and drive data integration into geospatial models. Sharing processing scripts (data-compilers) with data and model descriptions makes reusing and replicating models an easy task. In the case of LittoSIM-GEN, the data-compiler performs all required transformations to generate the LittoSIM data structure. However, not all data processing operations were automated. Topological verification, for example, is still done manually using GIS software.

The mapping approach presented here responds to the issue of genericity and reusability of LittoSIM and allows us to generalize the model (appendix A) to new territories where urban managers are concerned with the risks of coastal flooding (figure 11). Hence, reusing LittoSIM-GEN with other case studies is simplified and requires only three steps:

1. Gathering local data (shapefiles, DEM raster, bdy file) from available sources and databases.
2. Updating the structure (files and attribute names) of the XML mapping file to link it with the structure of the local source data.
3. Running the R processing script to generate the target LittoSIM structure.

In the French context where costal risk is managed at the inter-district level, there is a need to accompany each local community with the design of its strategic adaptation plan. The LittoSIM-GEN project participates in this effort by focusing on the integration of land use, hydrodynamic, and economic data, all combined in a learning tool for decisionmakers to encourage mid-long terms strategic adaptation to coastal risks. Learning tools such as LittoSIM-GEN need both to be adapted to the local context and to be rapidly applicable to the field. This is what the formalized data processing method presented in this paper allows.

To encourage third parties (such as local experts and technical offices) to apply LittoSIM-GEN in their territories, we suggest enhancing future versions with some new features. For example, we propose implementing a graphical interface to edit mappings between data sources and the LittoSIM structure, and then automatically generate the XML configuration file.

The proposed approach ensures the running of models but does not guarantee the validity of their outputs. Data transparency defines the application domain of the model and eases its replicability. However, the evaluation of models and their applicability to a specific context is still to be assessed by domain experts.

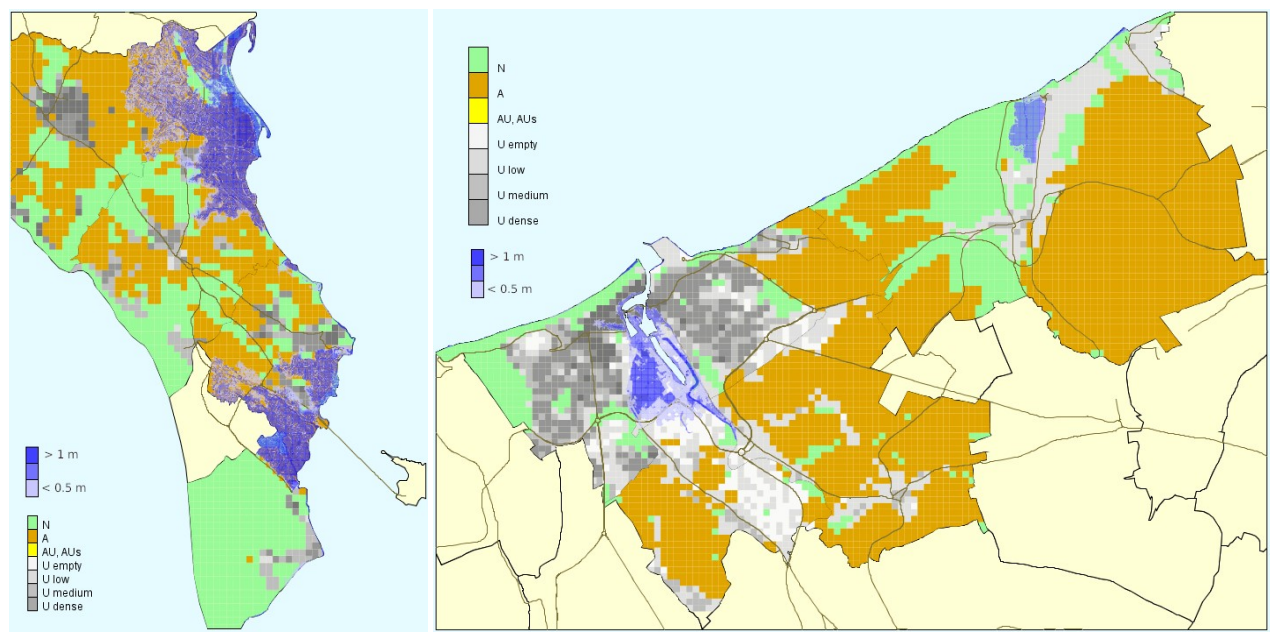

*(a) Oléron*                                    *(b) Normandie*

*Figure 11: LittoSIM-GEN in different territories. Multiple color legend refers to land use unit cells. The blue legend refers to water heights of submersion levels.*

# A LittoSIM-GEN

LittoSIM-GEN source code: *https://github.com/LittoSim/LittoSim_model/tree/LittoDev*

LittoSIM-GEN data-compiler: *https://github.com/LittoSim/LittoSim_model/wiki/LittoSIM-GEN-Data-compiler*

## Acknowledgments